\documentclass{article}

\pdfoutput=1
\usepackage{microtype}
\usepackage{graphicx}
\usepackage{subfig}
\usepackage[export]{adjustbox}
\usepackage{booktabs} 

\usepackage{hyperref}
\usepackage{amsmath,amssymb}

\newcommand{\sid}[1]{{\color{blue}{}}}

\newcommand{\jane}[1]{{\color{red}{}}}

\newcommand{\zeb}[1]{{\color{green}{}}}

\newcommand{\sam}[1]{{\color{blue}{}}}

\usepackage[accepted]{icml2018}
 
\icmltitlerunning{Meta-Learning with Episodic Recall}

\begin{document}

\twocolumn[
\icmltitle{Been There, Done That: Meta-Learning with Episodic Recall}

\begin{icmlauthorlist}
\icmlauthor{Samuel Ritter}{DM,Pton}
\icmlauthor{Jane X. Wang}{DM}
\icmlauthor{Zeb Kurth-Nelson}{DM,MPSUCL}
\icmlauthor{Siddhant M. Jayakumar}{DM}
\\
\icmlauthor{Charles Blundell}{DM}
\icmlauthor{Razvan Pascanu}{DM}
\icmlauthor{Matthew Botvinick}{DM,Gat}
\end{icmlauthorlist}

\icmlaffiliation{DM}{DeepMind, London, UK}
\icmlaffiliation{Pton}{Princeton Neuroscience Institute, Princeton, NJ}
\icmlaffiliation{MPSUCL}{MPS-UCL Centre for Computational Psychiatry, London, UK}
\icmlaffiliation{Gat}{Gatsby Computational Neuroscience Unit, UCL, London, UK}

\icmlcorrespondingauthor{Sam Ritter}{ritters@google.com}

\icmlkeywords{Meta-Learning, Episodic Memory, Deep Learning, Reinforcement Learning} 

\vskip 0.3in
]

\printAffiliationsAndNotice{} 

\begin{abstract}
Meta-learning agents excel at rapidly learning new tasks from open-ended task distributions; yet, they forget what they learn about each task as soon as the next begins. When tasks reoccur -- as they do in natural environments -- meta-learning agents must explore again instead of immediately exploiting previously discovered solutions. We propose a formalism for generating open-ended yet repetitious environments, then develop a meta-learning architecture for solving these environments. This architecture melds the standard LSTM working memory with a differentiable neural episodic memory. We explore the capabilities of agents with this \emph{episodic LSTM} in five meta-learning environments with reoccurring tasks, ranging from bandits to navigation and stochastic sequential decision problems. 
\end{abstract}

\section{Introduction}
Meta-learning refers to a process through which a learning agent improves the efficiency and effectiveness of its own learning processes through experience. First introduced as a core topic in AI research in the 1990s \cite{thrun1998learning, schmidhuber1996simple}, meta-learning has recently resurged as a front-line agenda item \citep{santoro2016one,andrychowicz2016learning,vinyals2016matching}. In addition to reviving the original topic, recent work has also added a new dimension by importing the theme of meta-learning into the realm of reinforcement learning \cite{wang2016learning,finn2017model,duan2016rl}. 

Meta-learning addresses a fundamental problem for real-world agents: How to cope with open-ended environments, which present the agent with an unbounded series of tasks. As such, meta-learning research has typically focused on the problem of efficient learning on new tasks, neglecting a second, equally important problem: What happens when a previously mastered task reoccurs? 
Ideally, in this case the learner should recognize the reoccurrence, retrieve the results of previous learning, and ``pick up where it left off.''  Remarkably, as we shall review, state-of-the-art meta-learning systems contain no mechanism to support this kind of recall. 

The problem of task reoccurrence is taken for granted in other areas of AI research, in particular work on life-long learning and continual learning \citep{Ring1995,kirkpatrick2017overcoming, thrun1996explanation}. Here, it is assumed that the environment rotates through a series of tasks, and one of the key challenges is to learn each task without forgetting what was learned about the others. However, work focusing on this problem has generally considered scenarios involving small sets of tasks, avoiding the more open-ended scenarios and the demand for few-shot learning that provide the focus in meta-learning research. 

Naturalistic environments concurrently pose both of the learning challenges we have touched on, confronting learners with (1) an open-ended series of related yet novel tasks, within which (2) previously encountered tasks identifiably reoccur (for related observations, see \citealp{anderson1990adaptive,o2009fragment}). In the present work, we formalize this dual learning problem, and propose 
an architecture which deals with both parts of it.

\section{Problem Formulation}

\begin{figure*}[!ht]
    \centering
    \includegraphics[width=.95\textwidth]{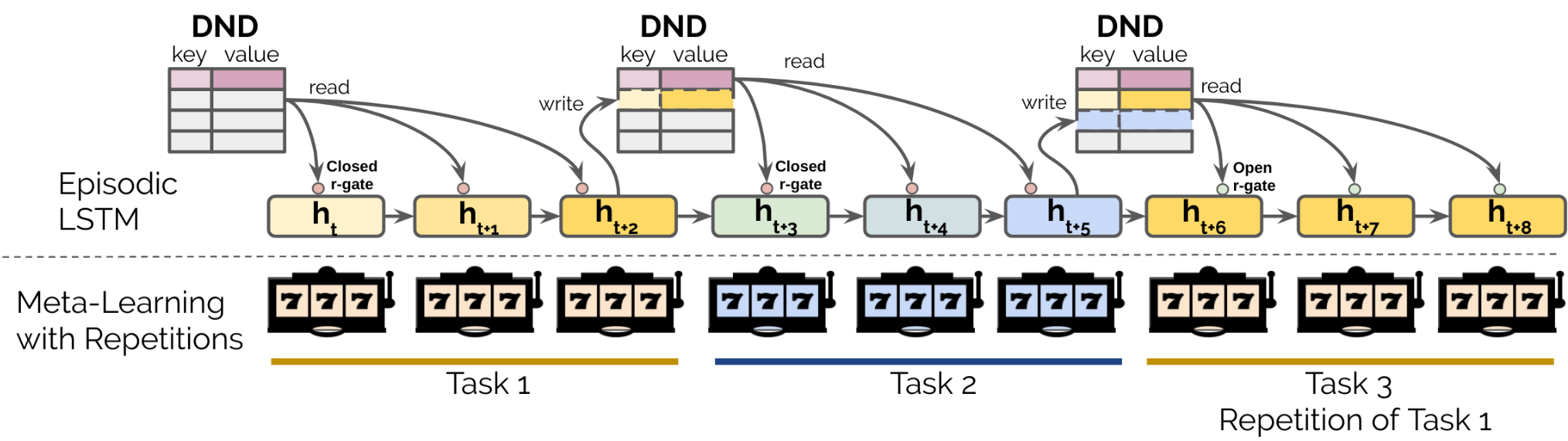}
    \caption{Model architecture and environment structure. Tasks, such as multi-armed bandits, are presented sequentially to the agent along with identifiable contexts, which are depicted here as colored bars. On each time step, the agent reads from long-term memory (DND) to retrieve cell states, which are reinstated through the multiplicative reinstatement gate (r-gate). At the end of each task, it writes its cell state to the DND.}
    \label{fig:arch_task}
\end{figure*}

Meta-learning research formalized the notion of an unbounded series of tasks by defining task distributions $\mathcal{D}$ over Markov Decision Processes (MDPs), then repeatedly sampling MDPs as $m\sim\mathcal{D}$ \cite{thrun1998learning}. To create open-ended sequences of novel and identifiably reoccuring tasks, we propose to instead sample MDPs $m$ along with contexts $c$ from stochastic task processes as follows
\begin{align}
(m_{n+1},c_{n+1}) | (m_1,c_1),...,(m_n,c_n) \sim \Omega(\theta, \mathcal{D}),
\end{align}
where $\Omega$ is a stochastic process with base distribution $\mathcal{D}$ and parameters $\theta$. $n$ indexes the tasks' positions in the sequence. From here on we will refer to these MDPs and their contexts, $(m_n,c_n)$, as the tasks $t_n$.

This framework can be used to challenge agents with precisely defined task reoccurrence frequencies. Consider for example a Blackwell-MacQueen urn scheme \citep{blackwell1973ferguson}, which samples MDP/context pairs $t$ successively as
\begin{align}
t_{n+1} | t_1,...,t_n \sim \frac{1}{\alpha + n} (\alpha \mathcal{D} + \sum_{i=1}^{n} \delta_{t_i}),
\end{align}
where $\delta_{t_i}$ is a point mass at $t_i$ and $\alpha$ is the concentration parameter. Intuitively, this scheme first samples an $(m,c)$ task pair from the distribution $\mathcal{D}$ then drops it into an urn. On the following and all subsequent steps, with probability $\frac{\alpha}{\alpha + n}$, the procedure samples a new task from $\mathcal{D}$ and drops it into the urn. Otherwise, it draws a task from the urn uniformly at random, copies it, and drops both the original and new copy into the urn \citep{teh2011dirichlet}. This process has a ``rich get richer'' property, whereby each occurrence of a task makes it more likely to reoccur, leading to Zipf-like distributions that are observed frequently in naturalistic environments (e.g., Huberman et al., \citeyear{huberman1998strong}). Task processes like this one may enable the development of agents that can cope with and ultimately take advantage of such important statistical properties of natural environments.

The remainder of this paper addresses the primary issue of identifying and reusing task solutions when the reoccurrence frequencies are uniform. Accordingly, the bulk of the experiments use the hypergeometric process, which repeatedly samples uniformly without replacement from a bag of tasks $S=\{t_1...t_{|S|}\}$ which contains duplicates of each task, so that $t_n \sim unif(S)$ \cite{terrell2006mathematical}. To solve this class of problems, we expect our deep reinforcement learning agents to: (1) meta-learn, capitalizing on shared structure to learn faster with each new task, and (2) avoid exploration when a task reoccurs, instead snapping back to hard-won effective policies.

\section{Agent Architecture}
\label{sec:agent_architecture}
We build on the \textit{learning to reinforcement learn} (L2RL) framework proposed by \citet{wang2016learning} and parallel work by \citet{duan2016rl}.
In L2RL, LSTM-based agents learn to explore novel tasks using inductive biases appropriate for the task distribution. They learn these exploration policies through training on tasks in which the reward on each time-step is supplied as an input, so that through training, the recurrent dynamics come to implement learning algorithms that use this reward signal to find the best policy for a new task. To execute such learning algorithms the LSTM must store relevant information from the recent history in its cell state.  
As a result, at the end of the agent's exposure to a task, the cell state contains the hard-won results of the agent's exploration.

Although this and the other meta-learning methods excel at acquiring knowledge of structure and then rapidly exploring new tasks, none are able to take advantage of task reoccurrence. In the case of meta-learning LSTMs, at the end of each exposure to a task, the agent resets its cell state to begin the next task, erasing the record of the results of its exploration.
To remedy this forgetting problem in L2RL, we propose a simple solution: add an episodic memory system that stores the cell state along with a contextual cue and reinstates that stored cell state when a similar cue is re-encountered later. This is inspired in part by evidence that human episodic memory retrieves past working memory states \citep{marr1971simple,hoskin2017refresh}.

To implement this proposal, we draw inspiration from recent memory architectures for RL. \citet{pritzel2017neural} proposed the differentiable neural dictionary (DND), which stores key/value pairs in each row of an array (see also \citealp{blundell2016model}). The values are retrieved based on k-nearest neighbor search over the keys, in a differentiable process that enables gradient-based training of a neural network that produces the keys. \citet{pritzel2017neural} achieved state-of-the-art sample efficiency on a suite of 57 Atari games by storing state-action value estimates as the DND's values and convolutional embeddings of the game pixels as the keys. Inspired by this success, we implement our cell state-based episodic memory as a DND that stores embeddings of task contexts $c$ as keys and stores LSTM cell states as values. In effect, this architecture provides context-based retrieval of the results of past exploration, addressing in principle the forgetting problem in L2RL.

However, an important design problem remains: how to incorporate the retrieved values with the ongoing L2RL process. The usual strategy for incorporating information retrieved from an external memory into its recurrent network controller is to pass the memories through parameterized transformations and provide the results as additional inputs to the network (e.g. \citealp{graves2016hybrid,weston2014memory}). While this has been effective in supervised settings, such transformations have proved difficult to learn by RL. Instead, we propose to make use of the unique relationship between the current working memory and retrieved states: not only do the retrieved states share the same dimensionality as the current working memory state, they also share the same \textit{semantics}. That is, the policy layer and the LSTM dynamics will operate in roughly the same way on axes of the retrieved states as they do on the axes of the current states. This affords the possibility of \textit{reinstating} the retrieved states; that is, adding them directly to the current cell state instead of treating them as additional inputs.

Such a reinstatement approach could fail if the reinstated activations interfere with necessary information stored in the current working memory state. We observe that the LSTM already solves a similar problem: it prevents incoming inputs from interfering with stored information, and vice versa, using the input and forget gates. We extend this same gating solution to coordinate among these and our new contributor to working memory. More precisely, the LSTM cell update equation \cite{hochreiter2001learning}
\begin{align}
\mathbf{c}_t = \mathbf{i}_t\circ \mathbf{c}_{in} + \mathbf{f}_t\circ \mathbf{c}_{t-1}
\end{align}
uses multiplicative gates $\mathbf{i}_t$ and $\mathbf{f}_t$, defined as follows, to coordinate between the contributions of current working memory and incoming perceptual inputs $\mathbf{c}_{in}$
\begin{align}
\mathbf{i}_t  = \sigma(\mathbf{W}_{xi}\mathbf{x}_t + \mathbf{W}_{hi}\mathbf{h}_{t-1} + \mathbf{b}_i)
\end{align}
\begin{align}
\mathbf{f}_t  = \sigma(\mathbf{W}_{xf}\mathbf{x}_t + \mathbf{W}_{hf}\mathbf{h}_{t-1} + \mathbf{b}_f).
\end{align}

\begin{figure}[ht]
    \centering
    \includegraphics[width=.45\textwidth]{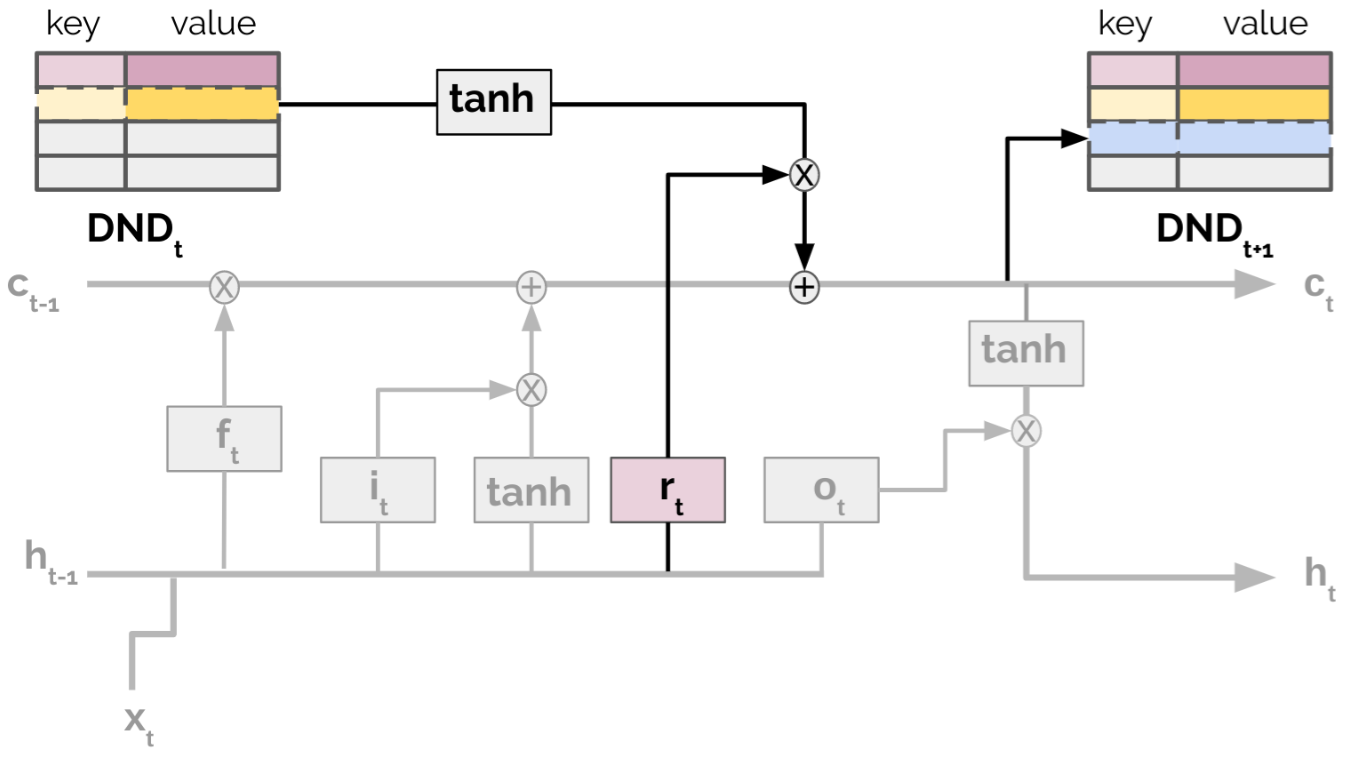}
    \caption{The episodic LSTM. In gray, the standard LSTM architecture. In bold, the proposed episodic memory and reinstatement pathways. See Section \ref{sec:agent_architecture} for details.}
    \label{fig:arch_detail}
\end{figure}

To the above update rule we add a new term for the contribution of reinstated working memory states,
\begin{align}
\mathbf{c}_t = \mathbf{i}_t\circ \mathbf{c}_{in} + \mathbf{f}_t\circ \mathbf{c}_{t-1} + \mathbf{r}_t\circ \mathbf{c}_{ep} 
\end{align}
\begin{align}
\mathbf{r}_t  = \sigma(\mathbf{W}_{xr}\mathbf{x}_t + \mathbf{W}_{hr}\mathbf{h}_{t-1} + \mathbf{b}_r),
\end{align}
where $\mathbf{r}_t$ is the reinstatement-gate, which, along with $\mathbf{i}_t$ and $\mathbf{f}_t$, coordinate among the three contributors to working memory. $\mathbf{c}_{ep}$ is the retrieved state from the episodic memory. Hereafter we will refer to this architecture as ``episodic LSTM'' (epLSTM, Figure \ref{fig:arch_detail}). 

\section{Experiments}
\label{Experiments}

We tested the capabilities of L2RL agents equipped with epLSTM (``epL2RL agents'') in five experiments. Experiments 1-3 use multi-armed bandits, first exploring the basic case where tasks reoccur in their entirety and are identified by exactly reoccurring contexts (Exp. 1), then, the more difficult challenge wherein contexts are drawn from Omniglot categories and vary in appearance with each reoccurrence (Exp. 2), and then, the more complex scenario where task components reoccur in arbitrary combinations (Exp. 3). Experiment 4 uses a water maze navigation task to assess epL2RL's ability to handle multi-state MDPs, and Experiment 5 uses a task from the neuroscience literature to examine the learning algorithms epL2RL learns to execute.

\begin{figure*}[ht]
   \centering
   \subfloat[\scriptsize{}]{
   \includegraphics[width=0.35\textwidth]{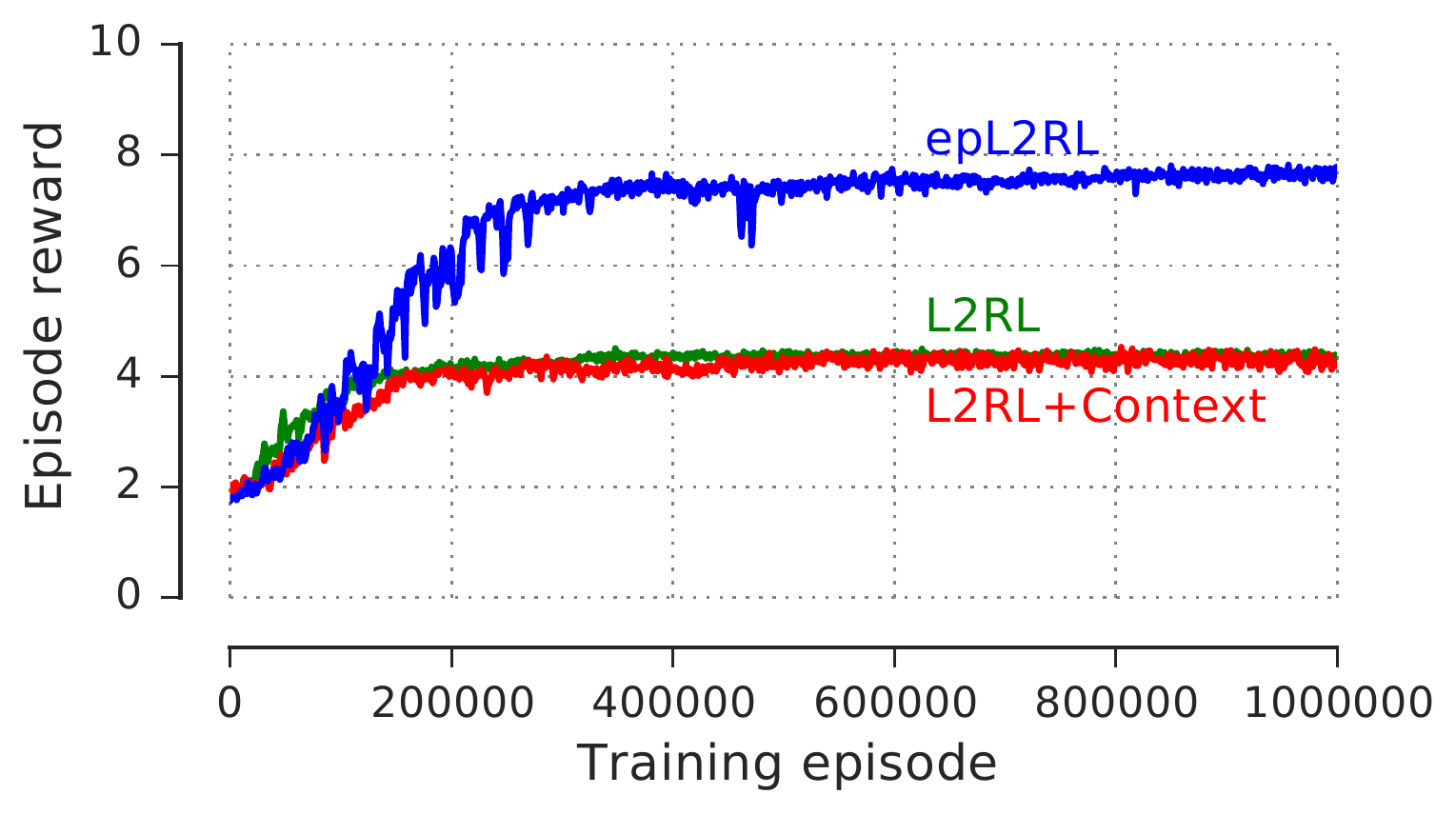}
   }
   \subfloat[\scriptsize{}]{
   \includegraphics[width=0.35\textwidth]{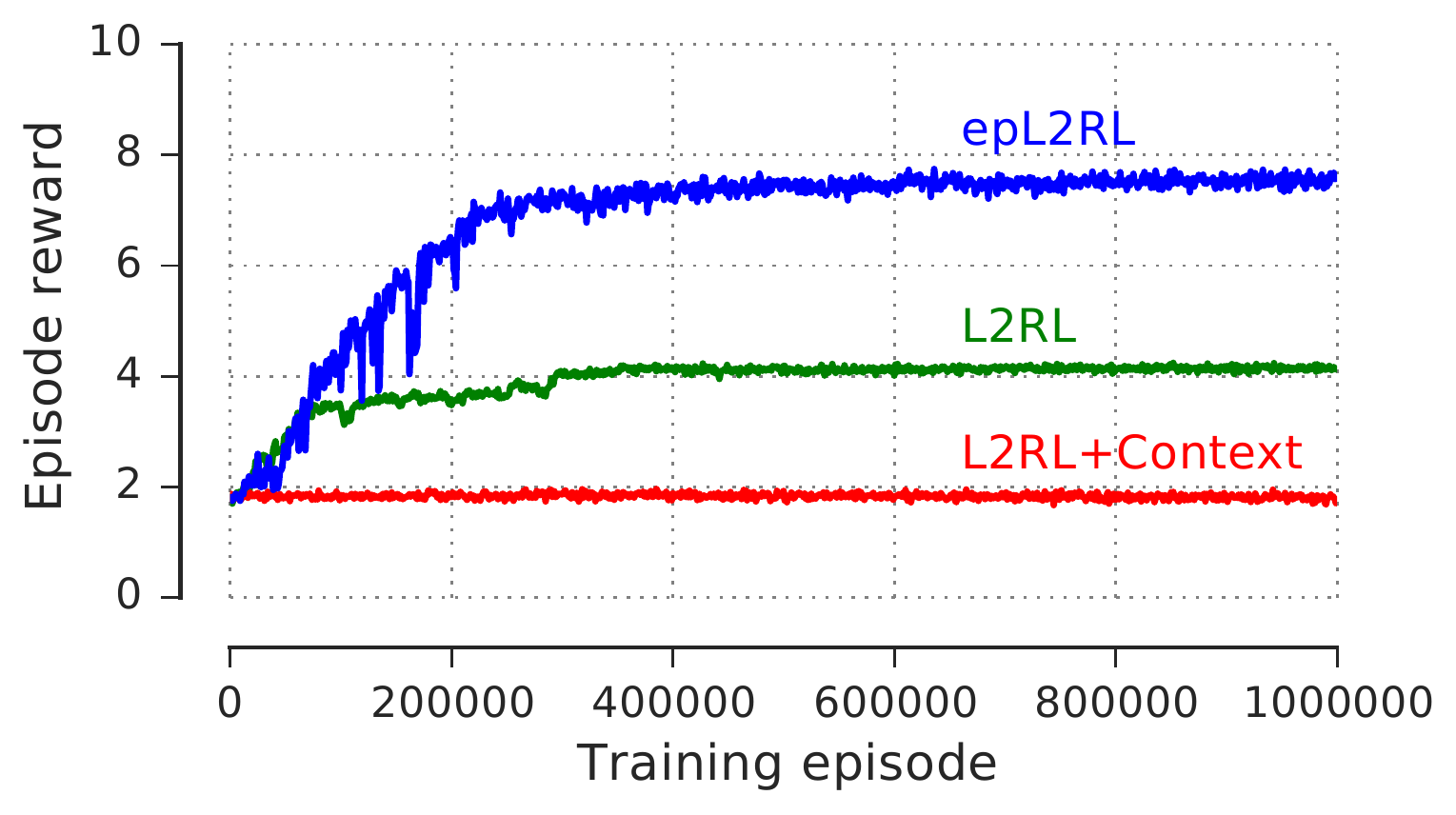}
   }
   
   \subfloat[\scriptsize{}]{
   \includegraphics[width=0.35\textwidth]{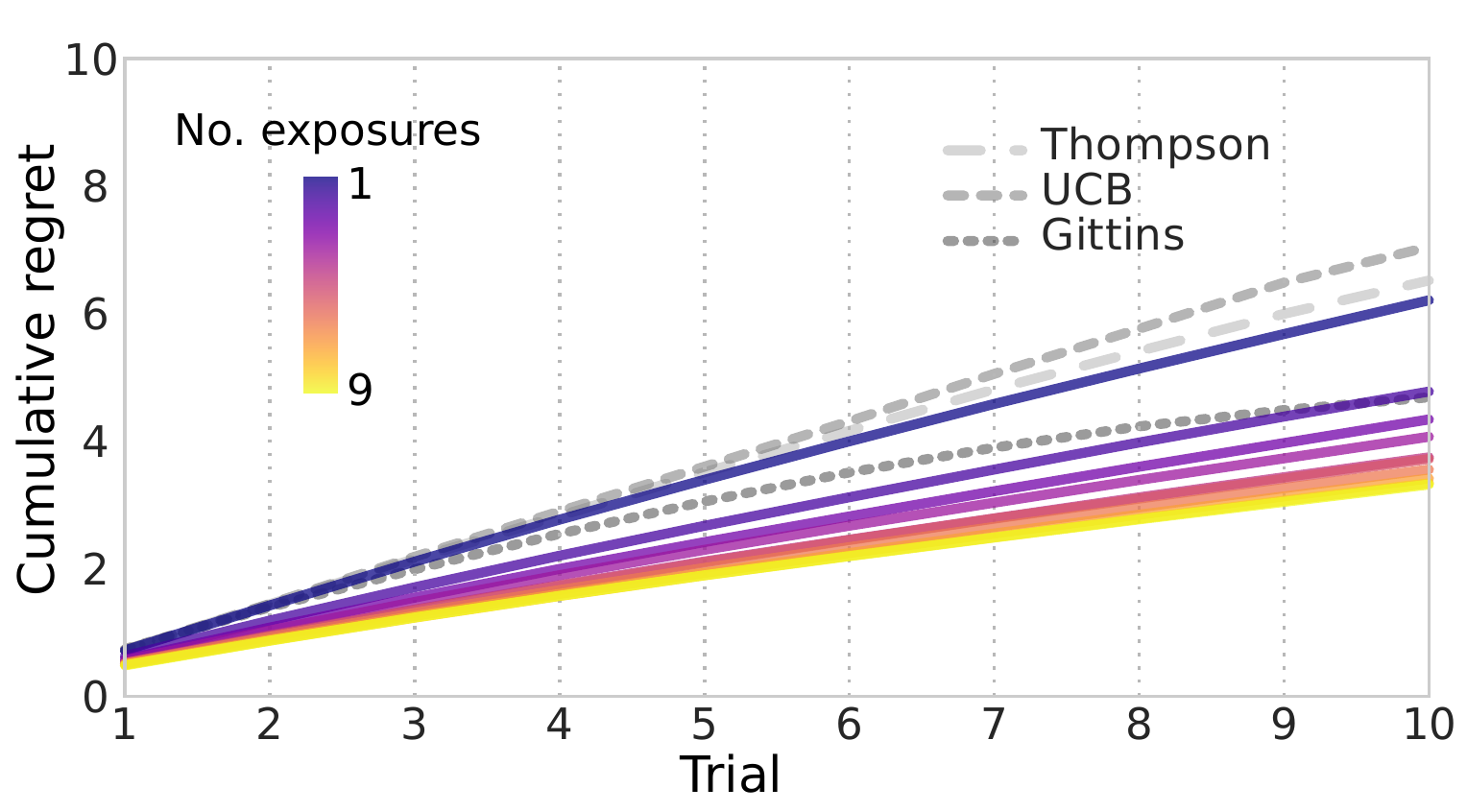}
   }
   \subfloat[\scriptsize{}]{
   \includegraphics[width=0.35\textwidth]{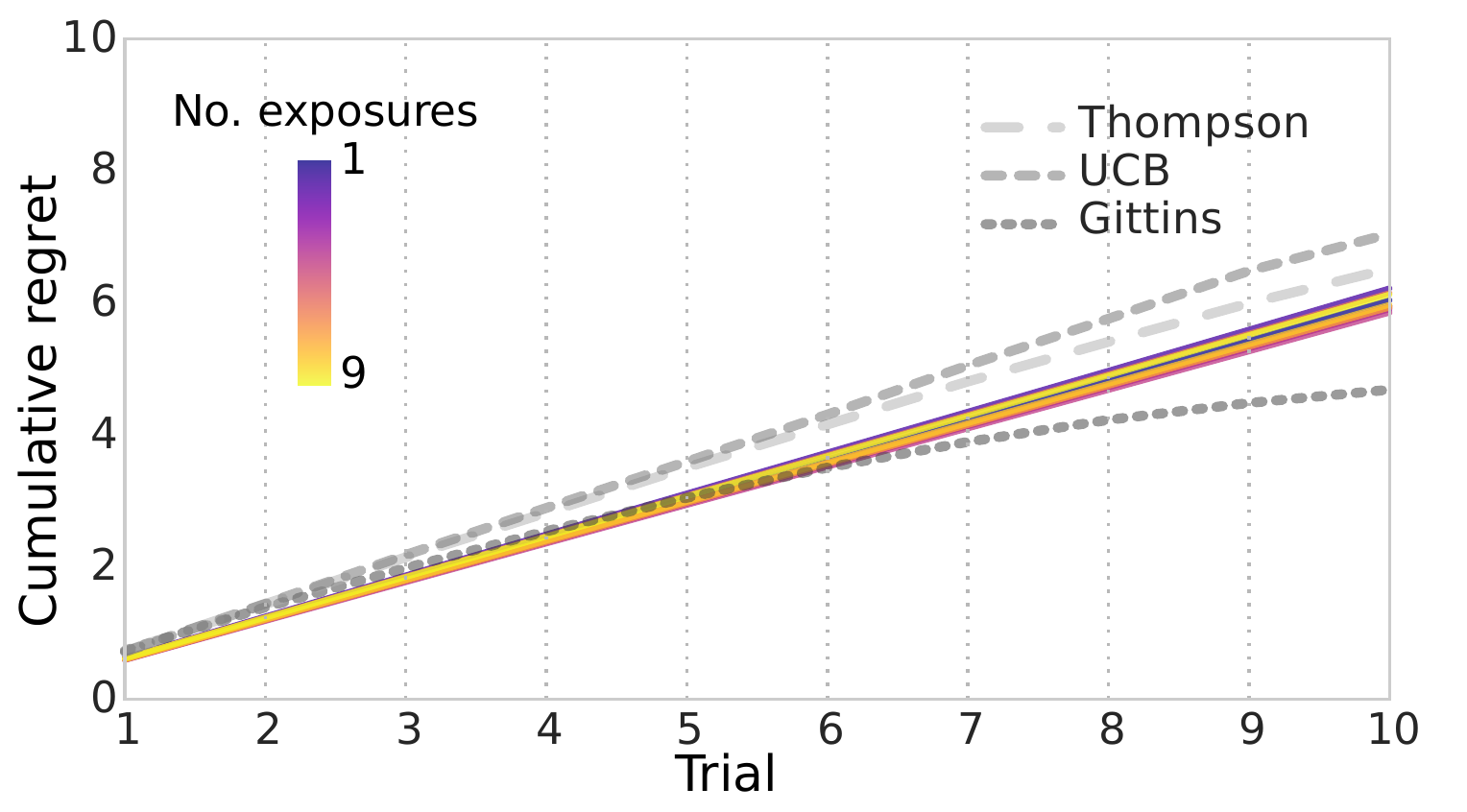}
   }
   \caption{Training and regret curves. (a) Barcode and (b) Omniglot bandit training curves, averaged over 3 runs. (c) Cumulative regret (expected loss in cumulative reward due to selecting suboptimal arms) for epL2RL agent on the Omniglot bandits task, averaged over evaluation episodes. Regret decreases as the number of exposures increases. The agent performs similarly to Gittins indices, Upper Confidence Bounds (UCB), and Thompson sampling on its first exposure, and outperforms them on subsequent exposures. (d) L2RL with no episodic memory performs on par with standard algorithms on Omniglot bandits, with no decrease in regret with more exposures.}
   
  \label{fig:barcode_omni} 
\end{figure*}

\subsection{Experiment 1: Barcode Bandits}
\label{sec:exp1_results}

Here we address a basic case of episodic repetition, where tasks reoccur in their entirety and are identified by exactly reoccurring contexts. The tasks in this experiment were contextual multi-armed bandits, elaborating the multi-armed bandits of \citet{wang2016learning} and \citet{duan2016rl}. Training consisted of episodes, in each of which the agent faced a set of actions (i.e., arms), where each arm yielded a reward with unknown probability. Each episode consisted of a series of pulls, throughout which the agent should efficiently find the best arm (explore) and pull it as many times as possible (exploit). During each episode, a context was presented which identified the reward probabilities. The challenge for the agent was to recall its past experience in a given context when the context reoccurred so that it could immediately exploit the best arm -- or continue its exploration where it left off.

The contexts in this experiment were binary strings, or ``barcodes'', so the full set of possible contexts was $C = \{0,1\}^l$, where $l=10$ was the barcode length. The reward parameters were structured such that all arms except for one had a reward probability of 0.1, while the remaining arm had a reward probability of 0.9. To prevent agents from overfitting the mapping between contexts and reward probabilities, we periodically shuffled this mapping throughout training. This shuffling serves a similar purpose to the randomization of bandit parameters between episodes: it forces the system to learn how to learn new mappings by disallowing it to overfit to a particular mapping. We hereafter refer to the sequences of episodes in which the mapping between context and bandit parameters is fixed as \textit{epochs}.

We sampled the sequence of tasks for each epoch as follows: we first sampled a set of unique contexts, and paired each element of that set randomly with one of the possible rewarding arm positions $b$, ensuring that each rewarding arm position was paired with at least one context. We then created a bag $S$ in which each $(c,b)$ pair was duplicated 10 times. Finally, we sampled the task sequence for the epoch by repeatedly sampling uniformly without replacement tasks $t_n = (c_n, b_n) \sim \textrm{unif}(S)$. There were 100 episodes per epoch and 10 unique contexts per epoch. Thus, each context was presented 10 times per epoch. There were 10 arms, and episodes were 10 trials long. 

In this and all following experiments, k was set equal to 1 in the k-nearest neighbors (kNN) search over the DND contents. Cosine distance and normalized inverse kernel were used for the kNN searches \citep{pritzel2017neural}. The DND was cleared at the end of each epoch, and the LSTM hidden state was reset at the end of every episode. Hyperparameters were tuned for the basic L2RL model, and held fixed for the other model variations.

Figure \ref{fig:barcode_omni}a shows training curves for the barcode bandits task. epL2RL widely outperforms the following baselines: L2RL, which is the agent from \cite{wang2016learning} and L2RL+Context, which is that same agent with the barcodes supplied as inputs to the LSTM. The results for this latter baseline indicate that epL2RL's performance cannot be matched by parametrically learning a mapping between barcodes and rewarding arms. epL2RL's increase in performance over L2RL suggests that the memory system is working, and in Experiment 2 we conducted further analysis to verify this hypothesis.

In this and all following experiments, we also tried a variant of the episodic memory architecture in which retrieved values from the DND, $c_{ep}$ were fed to the LSTM as inputs instead of added to the working memory through the r-gate. We found that feeding this 50 dimensional vector to the LSTM reduced performance to random chance. This is almost certainly because the large number of noisy additional inputs make it difficult to learn to use the previous action and previous reward. Because of its poor performance, we omit this variant from the plots\footnotemark. 

\footnotetext{We also tried shrinking the dimensionality of $c_{ep}$ using various linear layers and MLPs. We found that only with very aggressive compression (e.g. to $<$5 dimensions), it was possible to get this to work as well as gated reinstatement on only a subset of our tasks. This degree of compression is unlikely to be suitable for more complex tasks, so we omit further exploration of these architecture variants.}

\subsection{Experiment 2: Omniglot Bandits}
\label{sec:omni_bandits}
Tasks in the real world are rarely identifiable by exact labels; instead there is often an implicit classification problem to be solved in order to correctly recognize a context. 
To investigate whether epL2RL can handle this challenge, in Experiment 2 we posed an episodic contextual bandits task in which the contexts are Omniglot character classes \cite{lake2015human}. Each time the class reoccurs, the image shown to the agent will be a different drawing of the character, so that the agent must successfully search its memory for other instances of the class in order to make use of the results of its past exploration. The task sampling process, episode/epoch structure, and agent settings were the same as those described in Section \ref{sec:exp1_results}. 

We used pretrained Omniglot embeddings from \citet{kaiser2017learning}. This is a particularly appropriate method for pretraining because such a contrastive loss optimization procedure \cite{hadsell2006dimensionality} could be run online over the DND's contents, assuming some heuristic for determining neighbor status. Future work may try contrastive losses over the DND contents to learn perception online during RL; for discussion see Section \ref{sec:discussion}.

The training curves in Figure \ref{fig:barcode_omni}b show that epL2RL obtains more reward than its L2RL counterparts. L2RL+Context fails in this case because the Omniglot context embeddings are relatively large -- 128 dimensions -- drowning out the important reward and action input signals. In Figure \ref{fig:barcode_omni}c-d, we analyze the agents' behavior in more detail to understand how epL2RL is obtaining more reward. Figure \ref{fig:barcode_omni}c depicts epL2RL's cumulative regret, i.e. the difference between the expected reward of the optimal arm and the expected reward of the chosen arm, averaged over a set of evaluation episodes. The regret curves are split by the number of previous exposures to the Omniglot character class the agent experienced in that episode. From this we can see that during the first exposure, the agent accrues a relatively large amount of regret. On the second exposure it accrues significantly less, and this trend continues as the number of exposures increases. The possibility that this decrease in regret was due to gradient-based learning is ruled out by the fact that the weights were frozen during the evaluation episodes. This result indicates that the agent is able to recall the results of its previous exploration and to hone them further with each passing episode. In other words epL2RL is able to store partially completed solutions then later recall them to ``pick up where it left off''. 

In contrast, Figure \ref{fig:barcode_omni}d shows that no such improvement with increasing exposures occurs in the L2RL agent without episodic memory. Figure \ref{fig:barcode_omni}c-d compares agents' regret curves with those of traditional bandit algorithms: Gittins indices \citep{gittins1979bandit}, UCB \citep{auer2002finite} (which comes with theoretical finite-time regret guarantees), and Thompson sampling \citep{thompson1933likelihood}. We find that the L2RL agents compare favorably with these baselines as in \cite{wang2016learning} (Figure \ref{fig:barcode_omni}d). Further, we observe that epL2RL outperforms these algorithms after a few exposures (Figure \ref{fig:barcode_omni}c), suggesting that it is able to make use of the unfair advantage over these algorithms that its memory affords. 

\subsection{Experiment 3: Compositional Bandits}

Real world agents not only face tasks that reoccur in their entirety, but also task \textit{components} that reoccur in various combinations. This requires that agents compose memories of previously encountered task components. In this experiment we simulate this type of task composition. We continue to use the multi-armed bandit setup, but rather than associating a context with each set of arms, we associate a context with each arm individually. Each episode is then made up of an arbitrary combination of these context/arm pairs. When a context/arm reoccurs, it is not constrained to appear in the same ``position'' (i.e., it may not be associated with the same action) as when its appeared previously.

\begin{figure}[ht]
    \centering
    \subfloat[\scriptsize{}]{
    \includegraphics[width=0.35\textwidth]{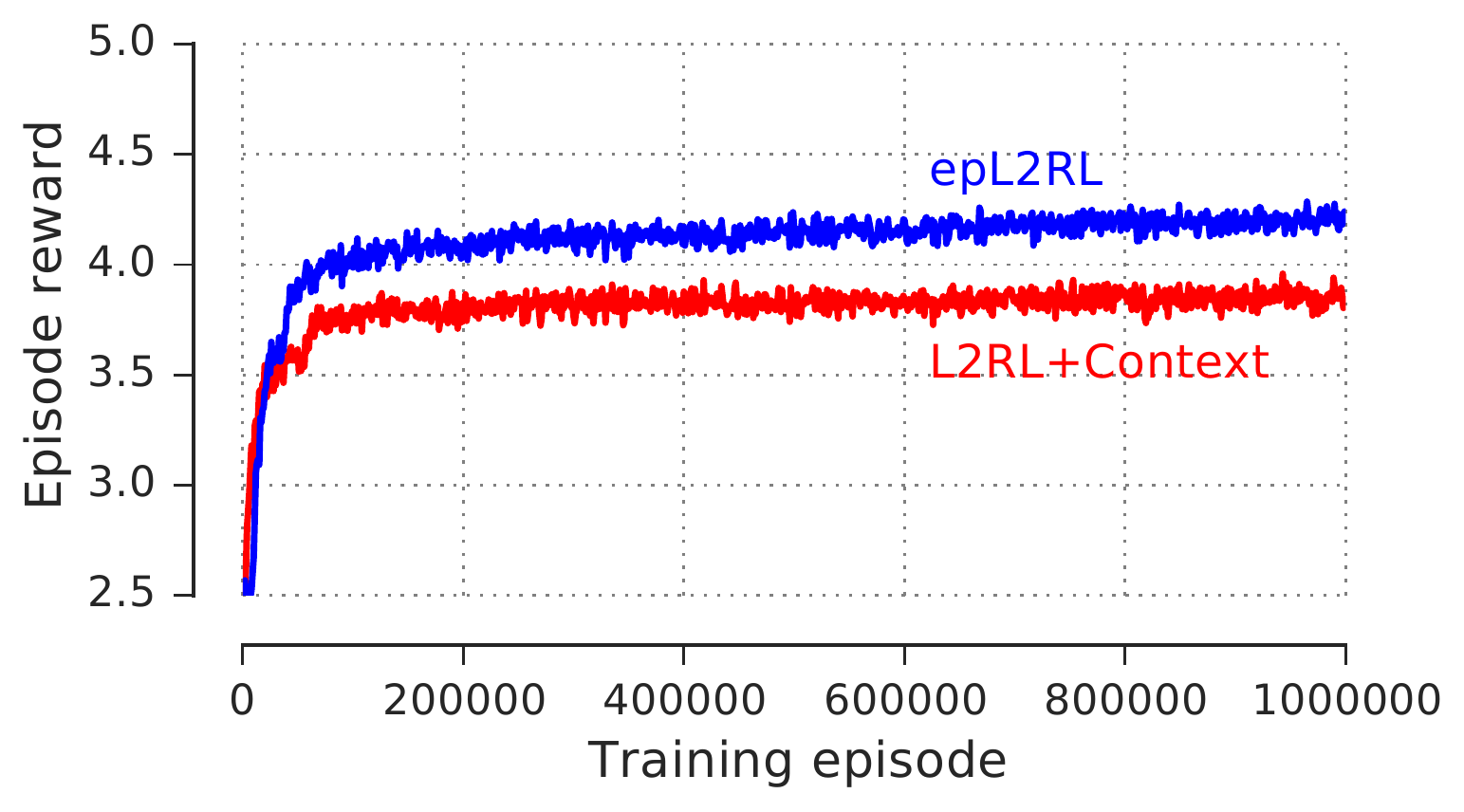}
    }
    
    \subfloat[\scriptsize{}]{
    \includegraphics[width=0.35\textwidth]{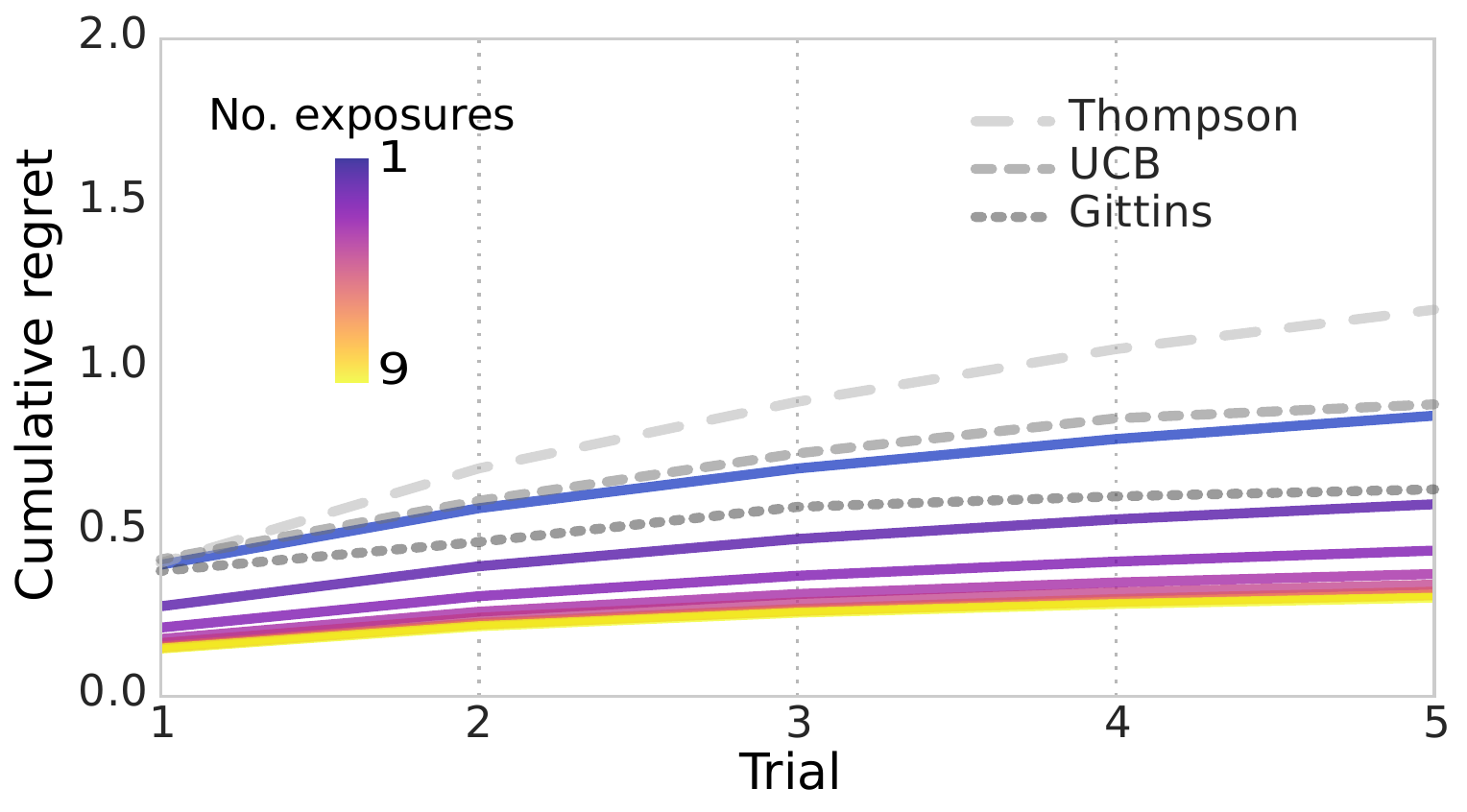}
    }
    \caption{(a) Training and (b) regret curves for compositional bandits. Training curves averaged over 3 runs, regret curves averaged over evaluation episodes.}
    \label{fig:comp_bandits}
\end{figure}

\begin{figure*}[ht]
    \centering
    \subfloat[\scriptsize{}]{
    \includegraphics[width=0.34\textwidth,valign=c]{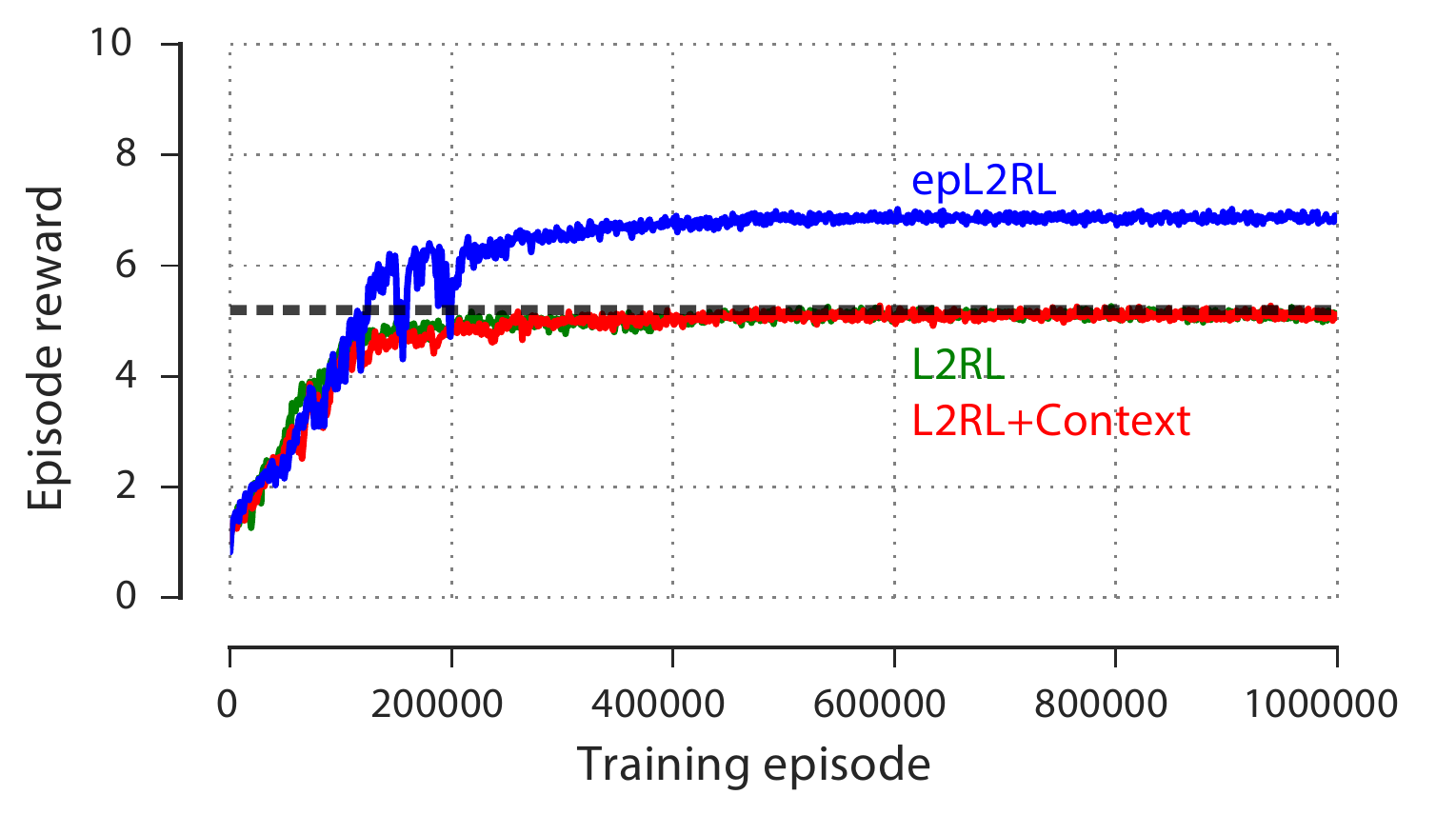}
    }
    \subfloat[\scriptsize{}]{
    \includegraphics[width=0.22\textwidth,valign=c]{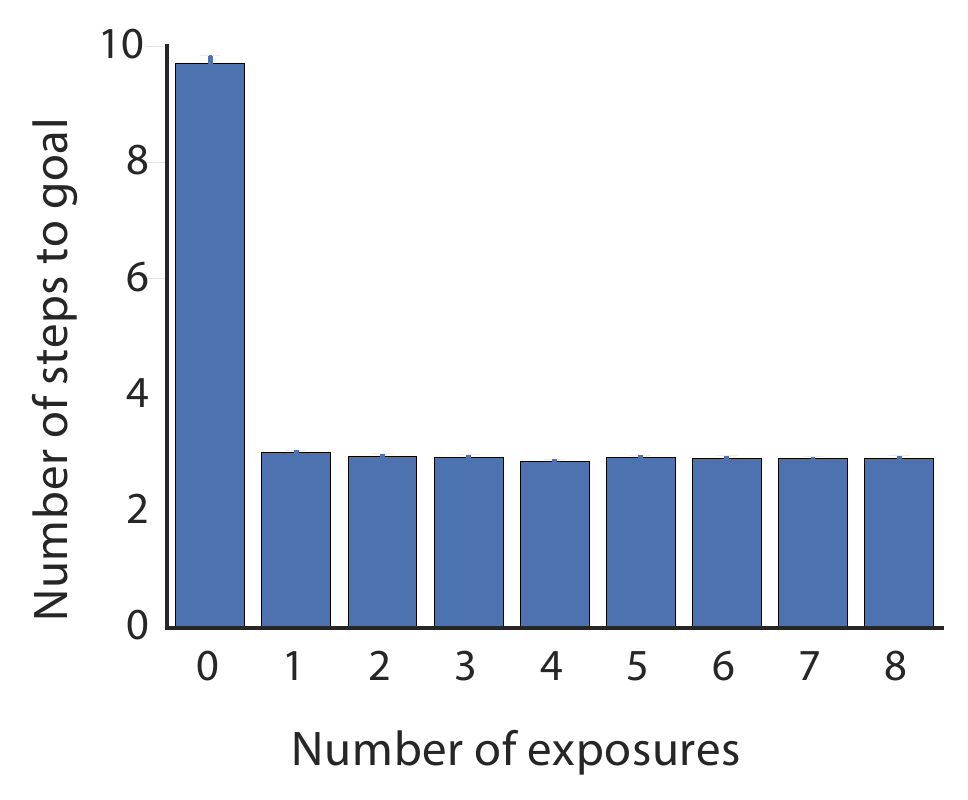}
    }
    \subfloat[\scriptsize{First exposure}]{
    \includegraphics[width=0.18\textwidth,valign=c]{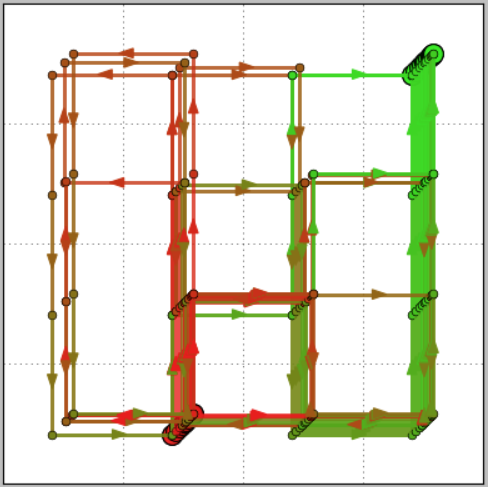}
    }
    \subfloat[\scriptsize{Second exposure}]{
    \includegraphics[width=0.18\textwidth,valign=c]{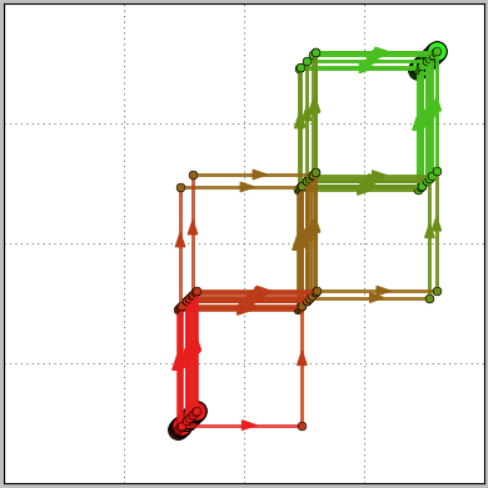}
    }
    \caption{Contextual Water Maze task. (a) Training curve averaged over 5 runs. The brown line indicates the maximum reward achievable without episodic memory. (b) Number of steps before the epL2RL agent reaches the goal after respawning, binned by number of previous exposures to the current context, data from evaluation episodes. After only one exposure, the agent can perform close to optimally. (c) Sample trajectories captured during evaluation for episodes with the same starting position and end goal during the first exposure to a context. (d) Same as c, but for the second exposure. Early steps are colored more red; later steps more green.}
    \label{fig:nav4x4}
\end{figure*}
This experiment used 2-armed bandits. The sampling process proceeded as follows: at the beginning of each epoch we first sampled a set $C_l$ of unique contexts, and paired each one with the low reward probability of 0.1, yielding pairs $(c,l)$. We then sampled a new set of contexts $C_h$ that contained no overlap with $C_l$, and paired each member of $C_h$ with the high reward probability of 0.9, yielding pairs $(c,h)$. 
Next we created a bag $S_h$ containing 5 duplicates of all \textit{high} rewarding contexts/reward probability pairs and a bag $S_l$ containing 5 duplicates of all \textit{low} rewarding contexts/reward probability pairs.
Finally, for each episode in the epoch we sampled randomly without replacement a low rewarding context $(c_n,l_n) \sim \textrm{unif}(S_l)$ and a high rewarding context $(c_n,h_n) \sim \textrm{unif}(S_h)$. The agent was then trained on an episode with these context/reward probability pairs.

In this setup the stimuli are paired directly with the reward probabilities, and the reward probabilities and stimuli together randomly swap positions on each trial. The stimuli are given to the LSTM as input in positions associated with the action, so that the LSTM must learn to discover the rewards associated with stimuli rather than with the arms. On each trial the epL2RL agent queries the DND using one of the two stimuli (selected randomly), retrieves a single cell state from the DND, then reinstates it as in the previous experiments. Otherwise, the agent settings were the same as those described in Section \ref{sec:exp1_results}. There were 1000 episodes per epoch and 400 classes per epoch, so that each class was shown 5 times. 

This setup and the aspect of the world that it models pose a specific problem for episodic memory. As in the previous experiments, the agent must retrieve holistic episodic memories based on context similarity; but, in this case the memory state vector the agent retrieves will contain not only information relevant to the task, but also information about another arm which is probably not present. Further, the information the memory contains about the relevant arm may refer to a trial in which that arm was shown in the other position. The agent must extract the relevant information from the memory without allowing the irrelevant information to affect its behavior, and must apply the correct information to the correct arm of the current task. In essence, because the environment state when the memory was saved differs from the current environment state, the agent must map information from the past onto the present. 

We find that epL2RL performs well even with no architectural modification, as evidenced by its increase in reward over L2RL+Context (Figure \ref{fig:comp_bandits}a). Further, the regret curves (Figure \ref{fig:comp_bandits}b) show that the epL2RL agent consistently reduces its regret after successive exposures to a given context. This decrease in regret cannot be attributed to gradient-based learning because the weights were frozen during the evaluation episodes. L2RL (without context) is omitted from Figure \ref{fig:comp_bandits}a because in this task it is impossible to determine the action-reward probabilities without the stimulus input. 

\subsection{Experiment 4: Contextual Water Maze}
In real-world episodic repetition, agents must explore stateful environments while managing contributions from episodic memory. The bandit tasks do not provide this challenge, so we now test our agents in minimal multi-state MDPs, specifically, contextual water mazes. In these tasks, the agent is shown a context barcode and must navigate a grid to find a goal using actions left, right, up, and down. The goal location is initially unknown. When the agent finds the goal, the agent's location is reset (at random), and, if it uses its LSTM working memory effectively, it can proceed directly back to the goal without needing to re-explore. In our setup the grid has 16 states (4x4) and the agent has 20 steps per episode to find the goal and return to it as many times as possible. 

The contextual (barcode) cues are mapped to goal locations in the same way they were mapped to rewarding arms in Experiments 1 and 2. Thus, once a goal is reached in a given context, the agent can use its memory of previous exploration to proceed directly to the goal without needing to explore again. The epoch structure and task sequence sampling process were the same as those described in Section \ref{sec:exp1_results}. The agent settings and architecture were also identical, except that (x,y) coordinates were provided as input. 

Figure \ref{fig:nav4x4}a shows that the epL2RL agent outperforms the L2RL baselines. It also significantly outperforms the maximum average reward that could be achieved without memory of past episodes. Figure \ref{fig:nav4x4}b shows the mean number of steps for the epL2RL agent to reach the goal from episode start, split by the number of previous exposures to the stimulus in the epoch. The means were calculated over a block of evaluation episodes (with learning rate set to zero). We see that the agent takes a much smaller number of steps to get to the goal after its first exposure. Since the data was recorded while the weights were frozen, this improvement cannot be explained by gradient-based learning, and thus must be the result of information stored in the episodic memory. 

Figures \ref{fig:nav4x4}c and \ref{fig:nav4x4}d show sets of epL2RL agent trajectories, again from episode blocks in which weights were frozen. Figure \ref{fig:nav4x4}c shows trajectories associated with the initial exposure to a context in an epoch, and a clear exploration policy is visible. Figure \ref{fig:nav4x4}d shows trajectories associated with the second exposure to the same context and starting location. These trajectories show that the agent in all cases navigated directly to the goal by one of the shortest paths. 

\subsection{Experiment 5: Episodic Two-Step Task}
\label{sec:twostep_task}

\citet{wang2016learning} used the two-step task \cite{daw2011model} to assess the degree to which L2RL learns to use model-based (MB) and model-free (MF) control. They found that L2RL learned to execute MB control, a remarkable result given that L2RL is trained via a purely model-free method. In our final experiment, we use a variant of the two-step task with episodic cues \cite{vikbladhCCN} to test whether our epL2RL agent can learn to execute \textit{episodic} MF and \textit{episodic} MB control.

In the classic two-step task, each episode\footnotemark consists of a single two stage MDP. On step 1, the agent can take one of two actions, $a_{1}$ or $a_{2}$ that will then lead through either a common transition or an uncommon transition to the resultant observable states $s_{1}$ or $s_{2}$. Rewards at $s_{1}$ and $s_{2}$ are drawn from a Bernoulli distribution which changes over time. In our setup, $[P(R|s_{1}), P(R|s_{2})]$ is either $[0.9, 0.1]$ or $[0.1, 0.9]$, and these parameters have a 10\% chance of reversing on every episode. In the episodic version of the task, cues (implemented here as barcodes) are presented at the second stage of the two-step episode. On step 1, the agent will either encounter no cue (an uncued episode), or a cue which matches the second-step context of an earlier episode $e$ (cued episode). On cued episodes, if the agent reaches the same state ($s_{1}$ or $s_{2}$) as it reached in episode $e$, then it's guaranteed to receive the exact same reward it received on episode $e$. Otherwise, it receives a draw from the currently active Bernoulli distribution for that state.

\footnotetext{Typically each two-step MDP traversal is referred to as a ``trial'' \cite{daw2011model}; we here use the term ``episode'' for consistency with Experiments 1-4.}

\begin{figure*}[ht]
    \centering
    \subfloat[\scriptsize{}]{
    \includegraphics[width=0.3\textwidth]{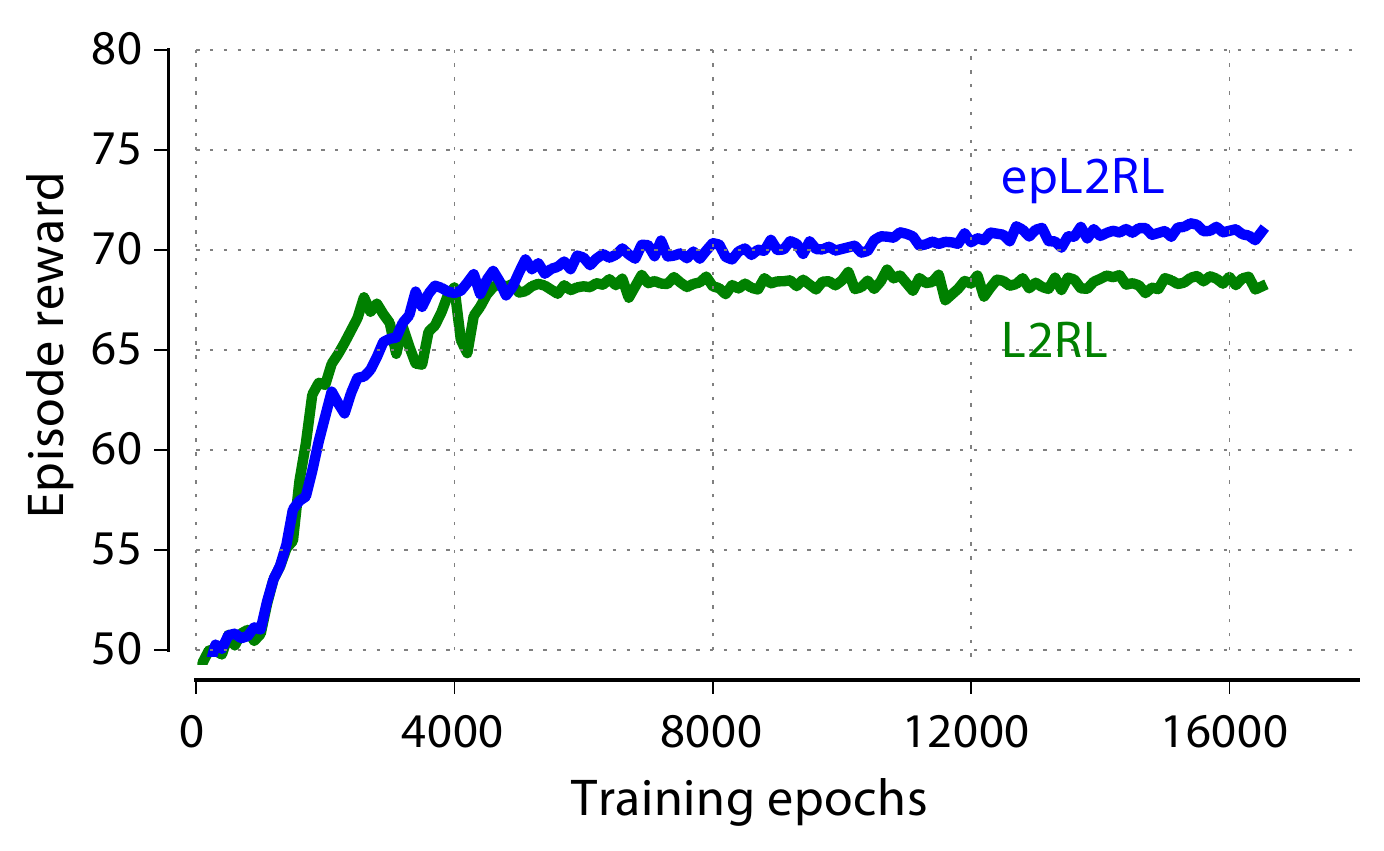}
    }
    \subfloat[\scriptsize{}]{
    \includegraphics[width=0.37\textwidth]{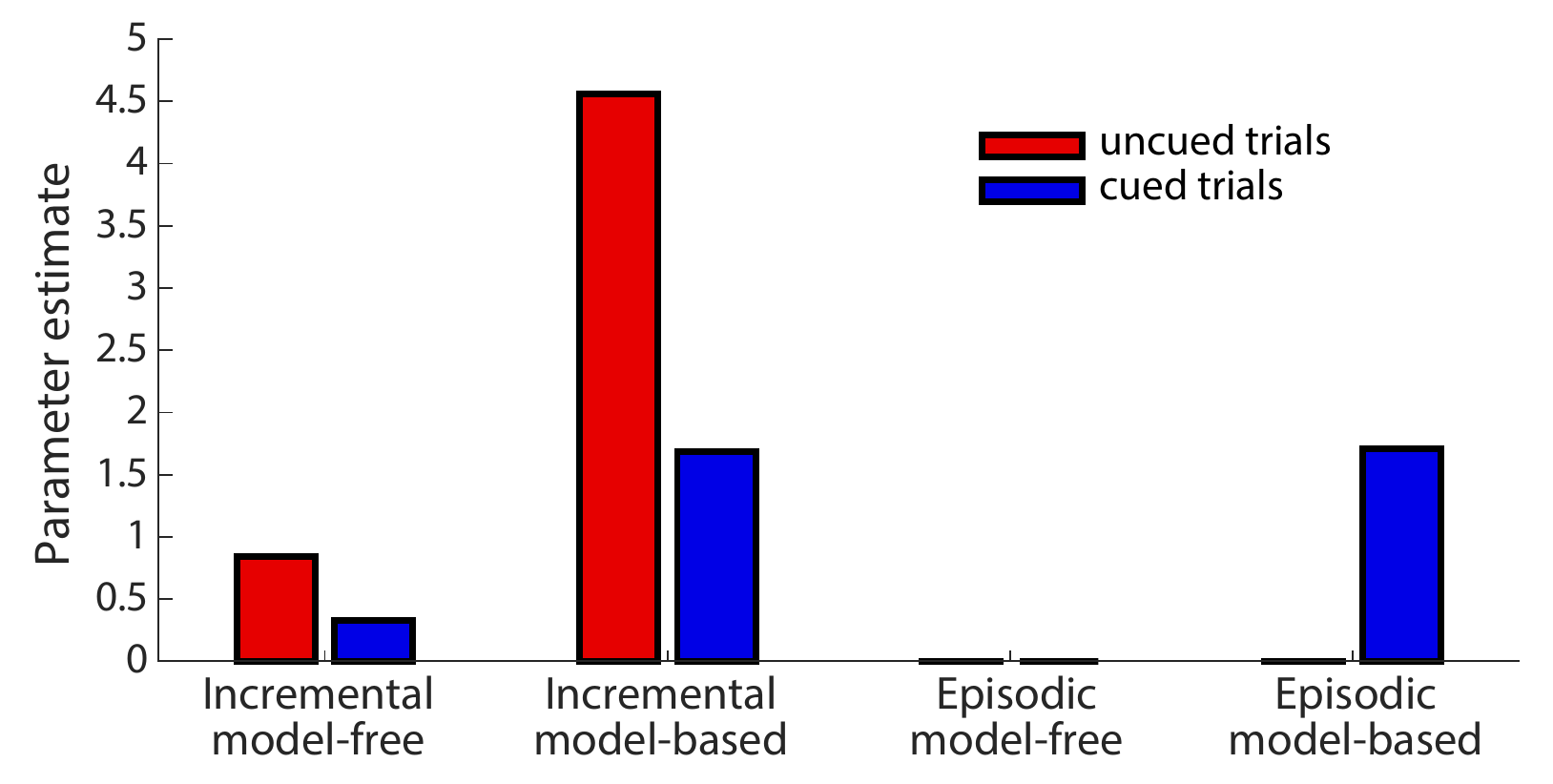}
    }
    \subfloat[\scriptsize{}]{
    \includegraphics[width=0.3\textwidth]{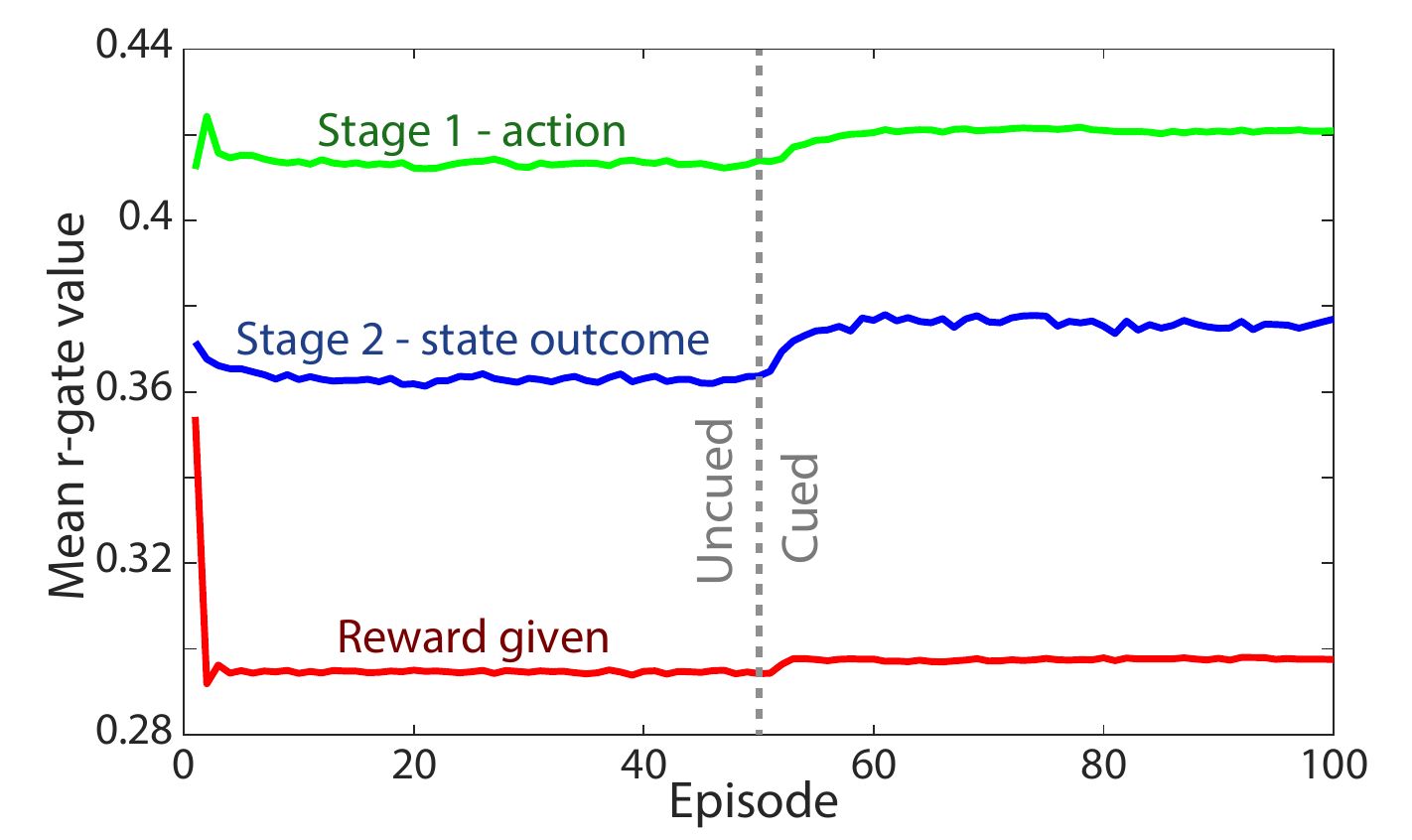}
    }
    \caption{Episodic two-step task results and analysis. See Section \ref{sec:twostep_task} for task description. (a) Training curves averaged over 10 runs show that epL2RL obtains higher reward than L2RL, providing evidence that the epL2RL agent is able to exploit the task's episodic cues. (b) Model fitting and parameter estimates for the different decision systems show that epL2RL uses IMF and IMB in the uncued episodes and IMF, IMB, and EMB on the cued episodes. (c) Time course of the mean r-gate values averaged over 500 epochs show that the gate is most open at the action stage, and is more open during cued episodes relative to uncued trials.}
    \label{fig:twostep}
\end{figure*}

This task is amenable to four valuation strategies; in other words, there are four different learning strategies epL2RL could learn to execute. The first is incremental model-free (IMF), wherein the agent takes the same action it took on the immediately previous episode if that episode was rewarded. The second is incremental model-based (IMB), whereby it takes the same action it took on the previous episode, only if that episode took a common transition. Agents with episodic memory could learn two additional strategies: episodic model-free (EMF) and episodic model-based (EMB), which operate like their respective incremental counterparts, but with respect to the past episode with which the episodic cue was associated. Beyond providing another setting in which to test the effectiveness of the reinstatement based episodic memory, this task also enables us to test specifically whether this system can learn to follow the more sophisticated and effective, but in principle more difficult to learn, episodic model-based behavior. Epochs consisted of 100 episodes each. The first half (50 episodes) of all epochs were uncued and the second half were cued. The agent settings were the same as those described in Section \ref{sec:exp1_results}.

We found that the epL2RL agent achieved more reward than the L2RL agent (Figure \ref{fig:twostep}a). To determine which algorithms epL2RL learned to use, we fit a choice model to the epL2RL agent's behavior. This model describes the probability of action $a_{1}$ as the softmax of a weighted sum of action values derived from IMF, IMB, EMF, and EMB control. We estimated these weights by maximum likelihood. Results showed that epL2RL does in fact learn to use an EMB policy, which it executes in tandem with both incremental learning strategies (Figure \ref{fig:twostep}b). Intriguingly, this is the same pattern of learning behavior observed in humans by \citet{vikbladhCCN}. For neuroscience implications, further analysis, and details see \citet{rittercogsci}.

\subsubsection{Analysis of the r-gates}
Next, we used this task to analyze the role of the epLSTM r-gates. Figure \ref{fig:twostep}c shows the mean r-gate value over the epoch, averaged over a block of evaluation epochs. These time courses are split by which stage of the two-step episode the agent was in. In all stages the r-gates open more when the cued episodes block starts. This makes sense, because in the uncued episodes there is no utility in reading from the memory. Further, we find that the r-gates were reliably more open on cued episodes in which the agent selected the correct action (mean r-gate value=0.365) than in cued episodes in which the agent selected the wrong action (mean r-gate value=0.358; two-tailed t-test p$<$1e-20). These observations provide preliminary evidence that the r-gates may work by allowing information from the DND into working memory when it is useful and gating it out when it is not. However, while the results discussed are highly statistically significant, the absolute magnitudes of the differences are very small. This suggests that other processes may be at work in governing the interplay between working memory, reinstated activations from the DND, and input in the epLSTM. Future work will be needed to explore this topic further.

\section{Discussion}
\label{sec:discussion}

This study constitutes a first step towards deep RL agents that exploit both structure and repetition in their environments. We introduced both a meta-learning training regime and a novel episodic memory architecture. Over the course of five experiments, we found our agents capable of recalling previously discovered policies, retrieving memories using category examples, handling compositional tasks, reinstating memories while traversing multi-state MDPs, and discovering the episodic learning algorithm humans use in a neuroscience-inspired task.

These results pave the way for future work to tackle additional challenges posed by real-world tasks. First, tasks often will not provide contexts as easily identifiable as barcodes, pretrained context embeddings will not be available, and contexts will be supplied in the same channel as other inputs. As such, a critical next step will be to learn to produce query keys. Two complementary approaches arise naturally for the epLSTM. First, the DND provides a mechanism for passing gradients through the retrieval process \cite{pritzel2017neural}; future work should explore the possibilities of using this learning pathway for epL2RL. Second, the contents of the DND provide an exciting opportunity for auxiliary training procedures, such as Kaiser et al.'s (\citeyear{kaiser2017learning}) algorithm. This procedure iteratively fills an array with embedding/label pairs and trains the embedding network by applying triplet loss to the nearest neighbors of new examples. Because the epLSTM's DND already accumulates embeddings and retrieves nearest neighbors, the marginal computational cost of applying such a contrastive loss is relatively low. The only challenge is to define the neighborhood function \cite{hadsell2006dimensionality}, which might be done using heuristics such as temporal contiguity.
 
Next, in many tasks of interest, contexts will not be fully observable during a single timestep; instead, information must be aggregated over time to produce an effective query embedding. Consider for example a task in which an agent can identify a previously solved maze only by observing several of its corridors.
Using the LSTM cell state, or a function thereof, as the query key is an appealing prospect for handling this challenge. 
Finally, in the present experiments, episode boundaries were clearly defined so that the agent could simply save at the end of each episode. In the real-world, events are not cut so cleanly, and agents must decide when to save, or save on every step and decide when to forget. Future work may pursue i) heuristics such as saving when reward is received and ii) learning approaches that leverage curricula beginning with short episodes amenable backpropagation through the storage process.

As a final note, although we developed the epLSTM to solve the forgetting problem in L2RL, such a reinstatement-based episodic memory system may be useful in other RL settings and for sequence learning problems in general. Future work may explore the potential of the epLSTM and other reinstatement-based memory systems in these domains.

\section*{Acknowledgements}
We thank Iain Dunning who developed, with help from Max Jaderberg and Tim Green, the asynchronous RL codebase we used to train our agents. We also thank the numerous research scientists and research engineers at DeepMind who worked on that code's Tensorflow and Torch predecessors. Further, we thank Alex Pritzel, Adri\`a Puigdom\`enech Badia, Benigno Uria, and Jonathan Hunt for their work on the DND library we used in this work. Finally, we thank Nicolas Heess for insightful discussion, and especially for his suggestion of water maze tasks; Ian Osband for sharing his bandits expertise; and Jack Rae for valuable discussion.

\bibliography{example_paper}
\bibliographystyle{icml2018}

\end{document}